\documentclass[lettersize,journal]{IEEEtran}
\usepackage{amsmath,amssymb,amsfonts}
\usepackage{algorithmic}
\usepackage{algorithm}
\usepackage{array}
\usepackage[caption=false,font=normalsize,labelfont=sf,textfont=sf]{subfig}
\usepackage{textcomp}
\usepackage{stfloats}
\usepackage{url}
\usepackage{verbatim}
\usepackage{graphicx}
\usepackage{epstopdf}
\usepackage{cite}
\usepackage{dsfont,pifont,bm,bbm,mathrsfs,mathtools,nicefrac, makecell, multirow}
\usepackage{float}
\usepackage{overpic}
\usepackage{marvosym}
\usepackage{hyperref}
\usepackage{cleveref}
\crefname{figure}{Fig.}{figures}
\crefname{section}{Sec.}{sections}
\crefname{equation}{Eq.}{Eqs.}
\crefname{table}{Tab.}{Tabs.}
\usepackage{booktabs}
\usepackage{orcidlink}

\usepackage{times}
\usepackage{epsfig}
\usepackage{gensymb}
\usepackage{pbalance}
\usepackage{color}
\usepackage{threeparttable}
\usepackage{wrapfig}

\begin{document}

\title{Motion Capture from Inertial and Vision Sensors}

\author{Xiaodong Chen~\orcidlink{0009-0002-2662-2206}, Wu Liu~\orcidlink{0000-0003-1633-7575},~\IEEEmembership{Senior Member,~IEEE,} Qian Bao~\orcidlink{0000-0002-4425-5481}, Xinchen Liu~\orcidlink{0000-0003-4931-8821},~\IEEEmembership{Member,~IEEE} \\ Ruoli Dai~\orcidlink{0009-0001-2577-1325}, Yongdong Zhang~\orcidlink{0000-0003-0066-3448},~\IEEEmembership{Fellow,~IEEE} and Tao Mei~\orcidlink{0000-0002-5990-7307},~\IEEEmembership{Fellow,~IEEE}
\thanks{Manuscript received October 29, 2025; revised February 9, 2026. This work is supported by the National Key Research and Development Program of China (NO. 2024YFE0203200), the National Nature Science Foundation of China (NO. U24A20329), and the Science Fund for Creative Research Groups (NO. 62121002). The associate editor of this paper was Amit Singh.}
\thanks{Xiaodong Chen, Wu Liu, Yongdong Zhang are with the School of Information Science and Technology, University of Science and Technology of China, Hefei 230022, China. (Corresponding author: Wu Liu, e-mail: cxd1230@mail.ustc.edu.cn, liuwu@ustc.edu.cn).}
\thanks{Xinchen Liu is with the Visual Technology Department, JD Explore Academy, Beijing 100176, China. Ruoli Dai is with the Office of the CTO, Noitom Technology Ltd., Beijing 100043, China. Tao Mei is with the Office of the CEO, HiDream.ai, Beijing 100080, China.} 
}


\maketitle

\begin{abstract}
Human motion capture is the foundation for many computer vision and graphics tasks. 
While industrial motion capture systems with complex camera arrays or expensive wearable sensors have been widely adopted in movie and game production, consumer-affordable and easy-to-use solutions for personal applications are still far from mature.
To utilize a mixture of a monocular camera and very few inertial measurement units (IMUs) for accurate multi-modal human motion capture in daily life, we contribute \textbf{MINIONS} in this paper, a large-scale \underline{M}otion capture dataset collected from \underline{IN}ertial and vis\underline{ION} \underline{S}ensors.
MINIONS has several featured properties:
1) large scale of over five million frames and 400 minutes duration;
2) multi-modality data of IMUs signals and RGB videos labeled with joint positions, joint rotations, SMPL parameters, etc.;
3) a diverse set of 146 fine-grained single and interactive actions with textual descriptions.
With the proposed MINIONS dataset, we propose a \textbf{SparseNet} framework to capture human motion from IMUs and videos by discovering their supplementary features and exploring the possibilities of consumer-affordable motion capture using a monocular camera and very few IMUs.
The experiment results emphasize the unique advantages of inertial and vision sensors, showcasing the promise of consumer-affordable multi-modal motion capture and providing a valuable resource for further research and development.
\end{abstract}


\section{Introduction}

Human motion capture is the process of recording human movement represented by a sequence of 3D positions and rotations of mesh or joints of the human body~\cite{hm36:journals/pami/IonescuPOS14, MINIONS:journals/corr/abs-2407-16341, chen2024clam}.
Industrial motion capture systems have been widely applied in movie and game production, sports analysis, medical diagnosis, etc.
However, these systems usually consist of tens of synchronized cameras or a group of wearable sensors and specific signal receivers~\cite{url:imu, url:vicon, TotalCapture:conf/bmvc/TrumbleGMHC17}.
Despite their high accuracy for human motion capture, individual consumers can hardly afford the high cost and learn professional configurations.
Therefore, this paper investigates accurate multi-modal human motion capture with consumer-affordable devices and easy-to-use operations for daily applications~\cite{zheng2021trand, ma2024humannerf, chen2022maple, liu2014vehicle} like eXtended Reality (XR), live video streaming, etc.

\begin{figure*}[t]
\centering
\includegraphics[width=0.99\linewidth]{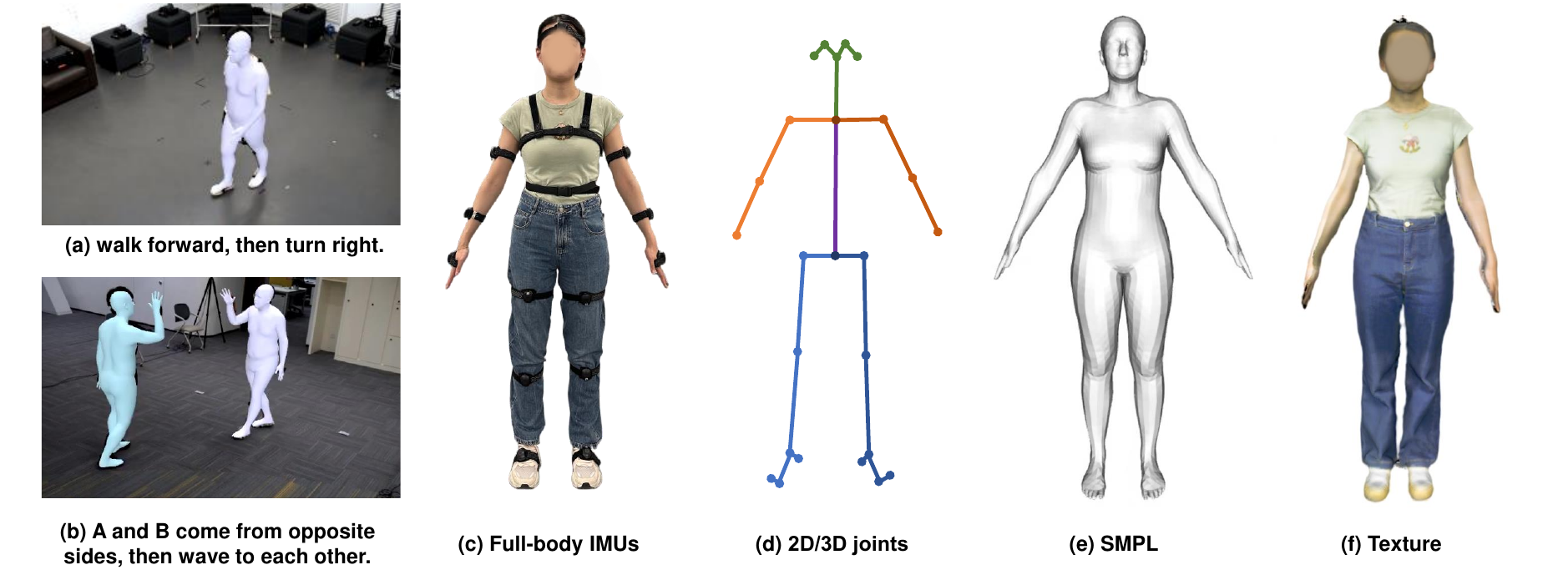}
\caption{\textbf{Overview of our MINIONS dataset.} It is collected by multiple types of sensors including eight 2K-resolution RGB cameras, Inertial Measurement Units (IMUs), and an RGB-D scanner. With the multi-modal data, we annotate human motion sequences with (d) 2D/3D joints, (e) the SMPL parameters, (f) the texture of each actor from a scanner, and fine-grained action types with textual descriptions.}
\vspace{-3mm}
 \label{fig1:overview}
\end{figure*}

Mainstream human motion capture solutions can be divided into marker-based and markerless systems.
The former usually requires actors to either wear tight clothes with reflective markers to be captured by an array of infrared cameras or bind a group of IMUs on their bodies that can be sensed by signal receivers~\cite{url:imu, url:vicon}.
They are expensive and inconvenient for actors to make natural movements. Specifically, IMU-based motion capture has an intrinsic limitation of global location drifting, which will limit long-term motions. 
The latter mainly consists of dozens of calibrated and synchronized cameras to surround the actors, which not only limits the scope of the actors but also needs complex configurations~\cite{hm36:journals/pami/IonescuPOS14}.
In addition, human motion capture from monocular RGB videos has also attracted the attention of researchers due to its cheapness and convenience~\cite{HMR:conf/cvpr/KanazawaBJM18,SMPL:journals/tog/LoperM0PB15}.
Some methods~\cite{smplify:conf/eccv/BogoKLG0B16, optimization:chen20233d} exploit the optimization-based pipeline that fits the motion parameters of Skinned Multi-Person Linear (SMPL)~\cite{SMPL:journals/tog/LoperM0PB15} models to human bodies in video frames.
Other approaches learn deep Convolutional Neural Networks (CNNs) on several benchmarks to regress the SMPL parameters of human bodies~\cite{ROMP:conf/iccv/SunBLFB021}.
However, they still suffer from depth ambiguity, occlusions, and quick motion instability in real-life scenarios.
Therefore, we aim to explore consumer-affordable multi-modal motion capture using a monocular camera and very few IMUs in daily life.

Despite the existing dataset TotalCapture~\cite{TotalCapture:conf/bmvc/TrumbleGMHC17} attempting to address these challenges, it suffers from the lack of variety in scenes, subjects, and actions, as well as the issue of the small scale. 
Furthermore, actors are required to wear tight-fitting attire throughout the data collection of TotalCapture, leading to inevitable distribution differences between daily life and the experimental environment. 
These weaknesses limit its potential for widespread application. 
Therefore, we attempt to build a large-scale dataset that covers diverse common actions performed by single or multiple subjects with daily clothes.
The datasets should also contain both videos and IMUs records with accurate human motion annotations like 3D position of joints, and SMPL parameters.

To this end, we contribute a large-scale \underline{M}otion capture dataset from \underline{IN}ertial and vis\underline{ION} \underline{S}ensors, named \textbf{MINIONS}, as shown in~\figurename~\ref{fig1:overview}.
The MINIONS dataset has several featured properties:
1) \textbf{Multi-modalities}: it is collected from eight 2K cameras and suits with 17 nine-axis IMU sensors. An RGB-D scanner is also adopted to obtain a textured mesh for each actor.
2) \textbf{Scalibility}: it contains over 5.5 million frames and 440 minutes of action sequences captured from different viewpoints.
3) \textbf{Diversity}: it covers 146 categories of fine-grained actions performed by 36 groups of actors (20 actors for single actions and 16 groups of actors for multi-person interactions).
4) \textbf{Abundance}: it provides abundant time-synchronized annotations for each frame, including 2D/3D joints, SMPL parameters~\cite{SMPL:journals/tog/LoperM0PB15}, fine-grained action class, and texture of actors.
We compare our MINIONS with existing motion capture datasets in~\tablename~\ref{tab1:overview}.

Based on MINIONS, we propose a two-branch framework, named~\textbf{SparseNet}, for motion capture by fusion of videos and IMUs.
The core of SparseNet is a deliberately designed two-branch structure grounded in Bayesian probability theory and Matrix Fisher priors~\cite{ta1:downs1972orientation, ta2:khatri1977mises}. 
For the sparse IMU signals, the Sparse IMUs branch takes inertial measurements as input to learn posture features, providing rotational probability distribution.
To overcome the incompleteness and location drift of sparse IMUs, we design a visual branch to supplement the missed joints and correct the location drift, estimating 3D bone probability distribution and pose probability prior.
This framework enables adaptive integration through a posterior fusion module that aligns distributions from both branches. 
Experiments reveal that the visual branch for monocular motion capture is often limited by jittering due to blurriness or occlusion, while the IMUs branch tends to suffer from global position drifts. 
However, a setup with four to six IMUs and a monocular camera can effectively achieve stable and virtually drift-free motion capture. 
Our experiment analysis offers substantial experimental evidence that paves the way for further research in this field.
Furthermore, MINIONS can also be a benchmark dataset for many other tasks such as 2D-to-3D pose estimation~\cite{Motionbert:zhu2023motionbert, video2dto3d:conf/cvpr/PavlloFGA19}, fine-grained action recognition~\cite{SlowFast:conf/iccv/Feichtenhofer0M19, TPN:conf/cvpr/YangXSDZ20}, gait activity assessment, etc.
The evaluation results of these tasks further exploit the potential of our dataset.

In summary, the contributions of this paper are three-fold:
\begin{itemize}
\item We build MINIONS, a large-scale human motion dataset from both RGB videos and IMUs, with multi-modal data, multiple actors, diverse actions, and rich annotations for the community.

\item We conduct SparseNet, a two-branch framework of multi-modal human motion capture using both inertial and vision sensors, and further explore the possibilities of consumer-affordable motion capture using a monocular camera and very few IMUs.

\item We provide extensive experiments on mainstream tasks, such as 2D-to-3D pose estimation, fine-grained action recognition, etc., opening up the potential of our datasets.
\end{itemize}

\section{Related Work}
\textbf{Motion Capture by Inertial Sensors}~\cite{imuonly1:conf/cvpr/YiZHSGT022, yi2022physical}.
IMUs-based human motion capture systems have been widely applied in the industry because they are robust to changes in illumination and occlusion.
Industrial solutions usually exploit full-body IMU sensors which must be bound to human bodies.
For example, the Perception Neuron system~\cite{url:imu} and the Xsens MVN system~\cite{Xsens:schepers2018xsens} both use 17 IMUs for human motion capture. 
However, the expensive cost, e.g., thousands of dollars for the IMU suit, and complex setup procedures before motion capture greatly raise the application threshold.
To reduce the cost of devices and setup, researchers study methods using very few IMUs such as six IMUs.
Recent approaches mainly adopted optimization-based or regression-based paradigms, which used machine learning or deep learning techniques.
For example, optimization-based methods~\cite{SIP:journals/cgf/MarcardRBP17, optimization:chen20233d} optimized all pose parameters in a sequence to find an optimal pose trajectory that is consistent with the acceleration parameters of sensors.
Regression-based methods~\cite{DIP:journals/tog/HuangKABHP18, transpose:journals/tog/YiZ021}, utilized deep neural networks to learn a mapping from sensor measurements to body positions and joint rotations from pair-wised data.
These learning-based methods provide a promising solution for sparse IMUs-based human motion capture. However, they still suffer from location drift due to the cumulative error of IMUs over a long time, e.g., several minutes.
Therefore, we consider visual information from RGB videos as supplementary cues for long-term stable motion capture.

\textbf{Motion Capture from Monocular Video}~\cite{imgonly1:conf/iccv/KocabasHHB21, imgonly2:journals/pami/ZhangTZLASL23, liu2022recent}.
Motion capture, i.e., 3D pose estimation, from monocular videos has been a popular topic in the computer vision community~\cite{liu2026lmagent}.
Recent approaches~\cite{tmm1:journals/tmm/GhafoorMB25, liu2024single, liu2025hoigen} can be divided into optimization-based methods and regression-based ones.
Optimization-based methods~\cite{smplify:conf/eccv/BogoKLG0B16, optimi1:conf/iccv/GuanWBB09, optimi2:conf/nips/SigalBB07} optimized parameters of the model, e.g., SMPL, to fit the joints or silhouettes of human bodies in video frames.
Regression-based methods, e.g., HMR~\cite{HMR:conf/cvpr/KanazawaBJM18} and ROMP~\cite{ROMP:conf/iccv/SunBLFB021}, learned deep neural networks from large-scale data to regress the model parameters given a single RGB image.
Although these methods have achieved excellent accuracy and real-time performance on these benchmarks, they may still generate poor results and temporal jitters due to occlusion, fast actions, and subtle movements in real-world scenarios.
Therefore, this paper aims to exploit sparse IMUs as supplementary cues to overcome these challenges.

\begin{table*}[t]
\small
\caption{\textbf{Comparisons of publicly available human motion datasets for motion capture.} MINIONS has more types of actions (Act), more frames, and a longer duration (minutes) of videos with single (S) and multiple (M) actors. Moreover, MINIONS provides both full-body IMU data and HD third-person view (TPV) RGB videos annotated with 3D SMPL Mesh of actors, joints~(KP), fine-grained action labels~(F-Act), and texture of actors (Texture).}
\centering
\begin{tabular}{l|c|cccc|cccccc} 
\midrule[1.5pt]
\multirow{2}{*}{Dataset} & \multirow{2}{*}{Year} & \multirow{2}{*}{Act} & \multirow{2}{*}{Frames} & \multirow{2}{*}{Duration} & \multirow{2}{*}{Actor} & \multicolumn{6}{c}{Modalities}        \\ 
\cline{7-12}
    &  &  &  &  &  & IMU & TPV-RGB & Mesh & KP & F-Act & Texture  \\ 
\midrule[1.5pt]
Human3.6M~\cite{hm36:journals/pami/IonescuPOS14} & 2014 & 17 & 3.6M & - & S & \textcolor[rgb]{0.6,0,0}{\ding{55}} & \textcolor[rgb]{0,0.6,0}{\ding{51}} & \textcolor[rgb]{0,0.6,0}{\ding{51}} & \textcolor[rgb]{0,0.6,0}{\ding{51}} & \textcolor[rgb]{0.6,0,0}{\ding{55}} & \textcolor[rgb]{0.6,0,0}{\ding{55}} \\

MPI-INF-3DHP~\cite{MPI3DHP:conf/3dim/MehtaRCFSXT17} & 2017 & 8 & 1.2M & - & S & \textcolor[rgb]{0.6,0,0}{\ding{55}} & \textcolor[rgb]{0,0.6,0}{\ding{51}} & \textcolor[rgb]{0.6,0,0}{\ding{55}} & \textcolor[rgb]{0,0.6,0}{\ding{51}} & \textcolor[rgb]{0.6,0,0}{\ding{55}} & \textcolor[rgb]{0.6,0,0}{\ding{55}} \\

TotalCapture~\cite{TotalCapture:conf/bmvc/TrumbleGMHC17} & 2017 & 5 & 1.9M & 50 & S & \textcolor[rgb]{0,0.6,0}{\ding{51}} & \textcolor[rgb]{0,0.6,0}{\ding{51}} & \textcolor[rgb]{0.6,0,0}{\ding{55}} & \textcolor[rgb]{0,0.6,0}{\ding{51}} & \textcolor[rgb]{0.6,0,0}{\ding{55}} & \textcolor[rgb]{0.6,0,0}{\ding{55}} \\

MuCo-3DHP~\cite{MuCo3DHP:conf/3dim/MehtaSMX0PT18} & 2018 & 8 & 0.5M & - & M & \textcolor[rgb]{0.6,0,0}{\ding{55}} & \textcolor[rgb]{0,0.6,0}{\ding{51}} & \textcolor[rgb]{0.6,0,0}{\ding{55}} & \textcolor[rgb]{0,0.6,0}{\ding{51}} & \textcolor[rgb]{0.6,0,0}{\ding{55}} & \textcolor[rgb]{0.6,0,0}{\ding{55}} \\

3DPW~\cite{3DPW:conf/eccv/MarcardHBRP18} & 2018 & - & 0.05M & - & M & \textcolor[rgb]{0,0.6,0}{\ding{51}} & \textcolor[rgb]{0,0.6,0}{\ding{51}} & \textcolor[rgb]{0,0.6,0}{\ding{51}} & \textcolor[rgb]{0,0.6,0}{\ding{51}} & \textcolor[rgb]{0.6,0,0}{\ding{55}} & \textcolor[rgb]{0.6,0,0}{\ding{55}} \\

DIP-IMU~\cite{DIP:journals/tog/HuangKABHP18} & 2018 & 64 & 0.3M & 92 & S & \textcolor[rgb]{0,0.6,0}{\ding{51}} & \textcolor[rgb]{0.6,0,0}{\ding{55}} & \textcolor[rgb]{0,0.6,0}{\ding{51}} & \textcolor[rgb]{0,0.6,0}{\ding{51}} & \textcolor[rgb]{0,0.6,0}{\ding{51}} & \textcolor[rgb]{0.6,0,0}{\ding{55}} \\ 

HUMAN4D~\cite{Human4d:chatzitofis2020human4d} & 2020 & 19 & 0.05M & - & M & \textcolor[rgb]{0.6,0,0}{\ding{55}}&\textcolor[rgb]{0,0.6,0}{\ding{51}}& \textcolor[rgb]{0.6,0,0}{\ding{55}}& \textcolor[rgb]{0,0.6,0}{\ding{51}}& \textcolor[rgb]{0.6,0,0}{\ding{55}}& \textcolor[rgb]{0,0.6,0}{\ding{51}} \\ 

RICH~\cite{RICH:huang2022capturing} & 2022 & - & 0.5M & - & M &  \textcolor[rgb]{0.6,0,0}{\ding{55}} & \textcolor[rgb]{0,0.6,0}{\ding{51}} & \textcolor[rgb]{0,0.6,0}{\ding{51}} & \textcolor[rgb]{0,0.6,0}{\ding{51}} & \textcolor[rgb]{0.6,0,0}{\ding{55}} & \textcolor[rgb]{0.6,0,0}{\ding{55}}  \\


H3WB~\cite{H3wb:zhu2023h3wb}& 2023 & 17 & 0.1M & - & S & \textcolor[rgb]{0.6,0,0}{\ding{55}} & \textcolor[rgb]{0,0.6,0}{\ding{51}} & \textcolor[rgb]{0,0.6,0}{\ding{51}} & \textcolor[rgb]{0,0.6,0}{\ding{51}} & \textcolor[rgb]{0.6,0,0}{\ding{55}} & \textcolor[rgb]{0.6,0,0}{\ding{55}} \\

RELI11D~\cite{yan2024reli11d}& 2024 & 15 & 0.2M & 199 & M & \textcolor[rgb]{0,0.6,0}{\ding{51}} & \textcolor[rgb]{0,0.6,0}{\ding{51}} & \textcolor[rgb]{0,0.6,0}{\ding{51}} & \textcolor[rgb]{0,0.6,0}{\ding{51}} & \textcolor[rgb]{0.6,0,0}{\ding{55}} & \textcolor[rgb]{0.6,0,0}{\ding{55}} \\

EMHI~\cite{fan2025emhi}& 2025 & 39  & 3.1M & - & S & \textcolor[rgb]{0,0.6,0}{\ding{51}} & \textcolor[rgb]{0.6,0,0}{\ding{55}} & \textcolor[rgb]{0,0.6,0}{\ding{51}} & \textcolor[rgb]{0,0.6,0}{\ding{51}} & \textcolor[rgb]{0.6,0,0}{\ding{55}} & \textcolor[rgb]{0.6,0,0}{\ding{55}} \\

\midrule
MINIONS (Ours) & - & \textbf{146} & \textbf{5.5M} & \textbf{440} & \textbf{M} & \textbf{\textcolor[rgb]{0,0.6,0}{\ding{51}}} & \textbf{\textcolor[rgb]{0,0.6,0}{\ding{51}}} & \textbf{\textcolor[rgb]{0,0.6,0}{\ding{51}}} & \textbf{\textcolor[rgb]{0,0.6,0}{\ding{51}}} & \textbf{\textcolor[rgb]{0,0.6,0}{\ding{51}}} & \textbf{\textcolor[rgb]{0,0.6,0}{\ding{51}}} \\
\midrule[1.5pt]
\end{tabular}

\label{tab1:overview} 
\end{table*}

\textbf{Motion Capture by Combination Schemes.}
Motion capture from both videos and IMUs has also attracted the attention of researchers.
Existing methods~\cite{fusion2:journals/tip/HenschelMR20, fusion5:conf/aaai/LiangHZLWYX23, fusion6:journals/tbe/ShinLH23} effectively improve the motion capture accuracy by eliminating multiplayer ambiguity in videos and minimization of location drift of IMUs.
For example, Gilbert~\textit{et al.}~\cite{fusion1:journals/ijcv/GilbertTMHC19} fused multi-channel volumetric data from multi-view cameras and IMU signals to estimate 3D joints.
They also built a dataset, TotalCapture~\cite{TotalCapture:conf/bmvc/TrumbleGMHC17}, as listed in~\tablename~\ref{tab1:overview}.
However, TotalCapture only contained 50-minute videos of five types of actions performed by five subjects, limiting the scalability of the dataset and related methods.
During the data collection of TotalCapture, actors are mandated to wear tight-fitting attire, resulting in unavoidable discrepancies between the attire worn in daily life and that of the experimental setting.
Therefore, we focus on building a large-scale motion capture dataset of daily life actions containing both RGB videos and IMUs and providing a more practical motion capture paradigm using monocular videos and sparse inertial signals for daily applications.

\section{The MINIONS Dataset}

The construction of our proposed MINIONS dataset consists of~\ref{dataset:setup} Multimedia Hardware Setup,~\ref{dataset:calibration} Calibration,~\ref{dataset:textured} Textured Mesh Reconstruction, ~\ref{dataset:joints} Global Motion Annotations, and ~\ref{subsec:Statistics} Dataset Statistics.

\subsection{Multimedia Hardware Setup}
\label{dataset:setup}
\begin{figure*}[t]
\centering
\includegraphics[width=0.95\textwidth]{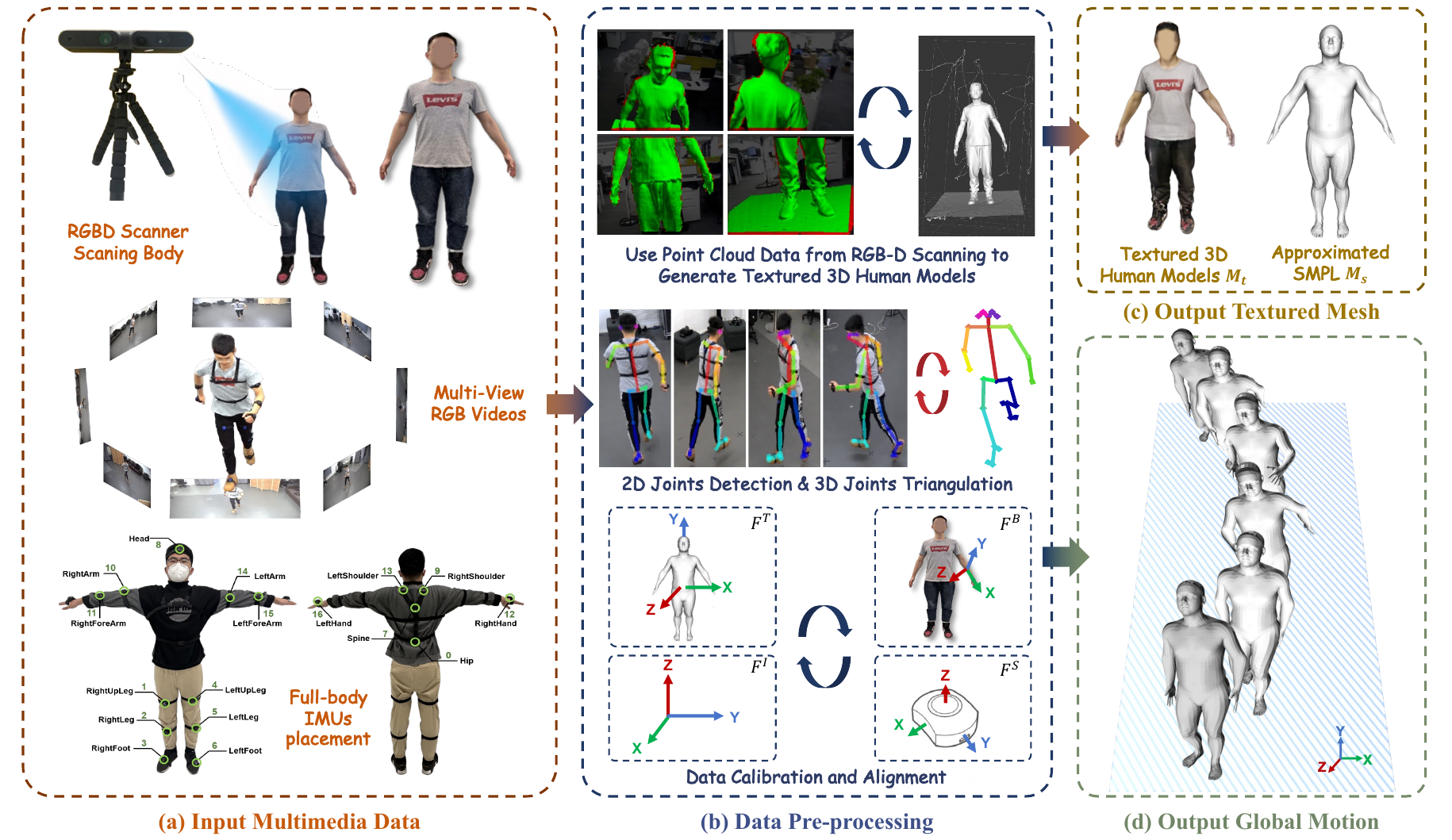} 
\caption{\textbf{Overview of Dataset Construction.} (a) Input multimedia data collected from multimedia hardware, including the RGB-D scanner, multiple synchronized cameras, and full-body IMU suits; 
(b) Data pre-processing, including point cloud data conversion, 2D joint detection, 3D joints triangulation, and IMUs data alignment; 
(c) Textured 3D human models and approximated SMPL shape; 
and (d) Global motion recovery from inertial and visual results.}
\label{fig2:datasetall}
\vspace{-3mm}
\end{figure*}

As shown in Fig.~\ref{fig2:datasetall} (a), we collect raw data in multiple scenes with multimedia hardware, including a professional RGB-D scanner, four to eight synchronized cameras, and full-body IMU suits with $17$ sensors.

\textbf{RGB-D Scanner:} A professional PUNE Scanner based on Microsoft's Kinect~\cite{url:xbox}, is used to obtain the human surface for texture and SMPL shape parameters recovery. 
The scanner can provide a scan accuracy of less than 1mm, with realistic texture maps of $1280\times 1024$ resolution.

\textbf{RGB Cameras:} High-speed industrial camera XiC MC023CG-SY~\cite{url:ximea} with a resolution of $1920\times 1200$ and 30fps, is used to capture multi-view RGB videos.
For synchronization, all cameras are connected with optical cables and triggered by a stabilized signal clock.

\textbf{IMUs:} The Perception Neuron Studio~\cite{url:imu} is applied to acquire full-body inertial data, including acceleration, angular velocity, and magnetic orientation.
Note that the synchronization between all cameras and IMUs is dependent on the calibrated timecode devices.

\subsection{Calibration}
\label{dataset:calibration}
\textbf{Calibration of Cameras.}
Following the general image-based calibration, we obtain coarse camera intrinsic matrix $K$ and external matrix $R|T$ according to Zhang's solution~\cite{zhang:journals/pami/Zhang00}. 
To improve the calibration quality, we manually mark the points on the ground to obtain a fine calibration. The average distance between calibration points and re-projected points is 0.63 pixels on a resolution of $1920\times1200$.

\textbf{Calibration of IMUs.}
The raw sensor data are measured in the IMU local coordinate system and should be aligned with the SMPL coordinate system. 
So, we transform the raw data following the DIP approach~\cite{DIP:journals/tog/HuangKABHP18}.
Through initial T-pose calibration, we obtain maps $R^{IS}$:$F^S\rightarrow{F^I}$, $R^{TI}$:$F^I\rightarrow{F^T}$ and $R^{TB}$:$F^B\rightarrow{F^T}$ between sensor coordinate system $F^S$, inertial coordinate system $F^I$, SMPL coordinate system $F^T$, and bones coordinate system $F^B$.
The sensor coordinate system $F^S$ is the local coordinate system attached to each IMU sensor, moving with the sensor.
The inertial coordinate system $F^I$ is a fixed global reference coordinate system, often aligned with gravity and magnetic north at the start of calibration.
The SMPL Coordinate System $F^T$ is the canonical coordinate system of the SMPL human body model, with its origin typically at the pelvis.
The Bones Coordinate System $F^B$ is the local coordinate system attached to an individual bone or joint, defining its local rotation.

\textbf{Calibration between Cameras and IMUs.} 
By aligning the T-pose skeleton in the camera coordinate system $F^C$ and the SMPL coordinate system $F^T$, we obtain the transformation between them $R^{TC}$:$F^C\rightarrow{F^T}$.

\subsection{Textured Mesh Reconstruction}
\label{dataset:textured}
As shown in the upper part of \figurename~\ref{fig2:datasetall} (b) and \figurename~\ref{fig2:datasetall} (c), we obtain RGB-D images from the scanner and convert them to point clouds to generate human models $M_t$ with textures in a canonical pose~\cite{texture2:conf/cvpr/PavlakosCGBOTB19, texture1:conf/cvpr/ZhangPBP17}. 
To obtain the static SMPL shape parameters $\beta$, we adjust the parameters $(\beta,\theta)$ of SMPL $M_s$ to be close to $M_t$ and in the interior of $M_t$ through the optimization as in~\cite{texture1:conf/cvpr/ZhangPBP17},
\begin{equation}
   E(\beta,\theta) = E_J + E_{skin} +  E_{cloth} +  E_{reg},
\label{equ1:texture}
\end{equation}
where $\beta$ and $\theta$ denote the shape and canonical pose parameters of SMPL, $E_J$ penalizes the error between the projection and the observed 2D joints, $E_{skin}$ keeps the scan points belonging to the skin close to the model $M_s$, $E_{cloth}$ prevents scan points belonging to clothes to be inside the model $M_s$, $E_{reg}$ is a priori term to make the results more reasonable.

\subsection{Global Motion Annotations}
\label{dataset:joints}
The global human motion annotations workflow includes 1) tracking; 2) 2D joint detection; 3) 3D joints triangulation, and 4) motion recovery from multimedia data. It provides high-quality annotations, including 2D and 3D joints, SMPL parameters, and person identity (ID). 
To ensure the quality of the dataset, each step is double-checked by experienced annotators, and the erroneous results of detection and tracking are adjusted manually.

\textbf{Tracking.}
Unlike the workflow of single-subject motion datasets~\cite{hm36:journals/pami/IonescuPOS14, MPI3DHP:conf/3dim/MehtaRCFSXT17, TotalCapture:conf/bmvc/TrumbleGMHC17}, our dataset track the 3D joints $\widetilde{P_{3d}}$ to get correct multi-person motion trajectories.
To achieve efficient and accurate tracking, we estimate the corresponding 3D boxes from the 3D joints $\widetilde{P_{3d}}$ and then match these boxes using the reid-based~\cite{he2023fastreid} StrongSORT algorithm~\cite{du2023strongsort}.
Each actor has a distinguished ID, and the erroneous tracking results (including ID switches and assigning new IDs) are adjusted with manual double-checking.

\textbf{2D Joints Detection.}
We adopt the HRNet-w48~\cite{HRNET:journals/pami/00010CJDZ0MTW0X21} as the 2D joints detector because of its excellent performance on the 2D joints detection benchmark~\cite{cocowhole1, cocowhole2}. 
After that, we post-process the 2D joints through DarkNet~\cite{DarkNet:conf/cvpr/ZhangZD0Z20} to reduce jitters and improve accuracy. 
The detection result contains 25 joints $P_{2d}$ of body, face, and feet in the same format as OpenPose~\cite{openpose}. 
We discard the uncertain joints with low confidence scores below $0.7$.

\textbf{3D Joints Triangulation.}
With the camera parameters $(K,R|T)$ and multi-view 2D joints $P_{2d}$, as shown in \figurename~\ref{fig2:datasetall} (b), we select top-$K$ views with the smallest reconstruction error to triangulate the initial 3D joints $\widetilde{P_{3d}}$, instead of all redundant camera views. 
To guarantee the quality of reconstruction, we discard the views with large reconstruction errors higher than $0.01$ meters.

\begin{figure}[t]
\centering
\includegraphics[width=0.99\linewidth]{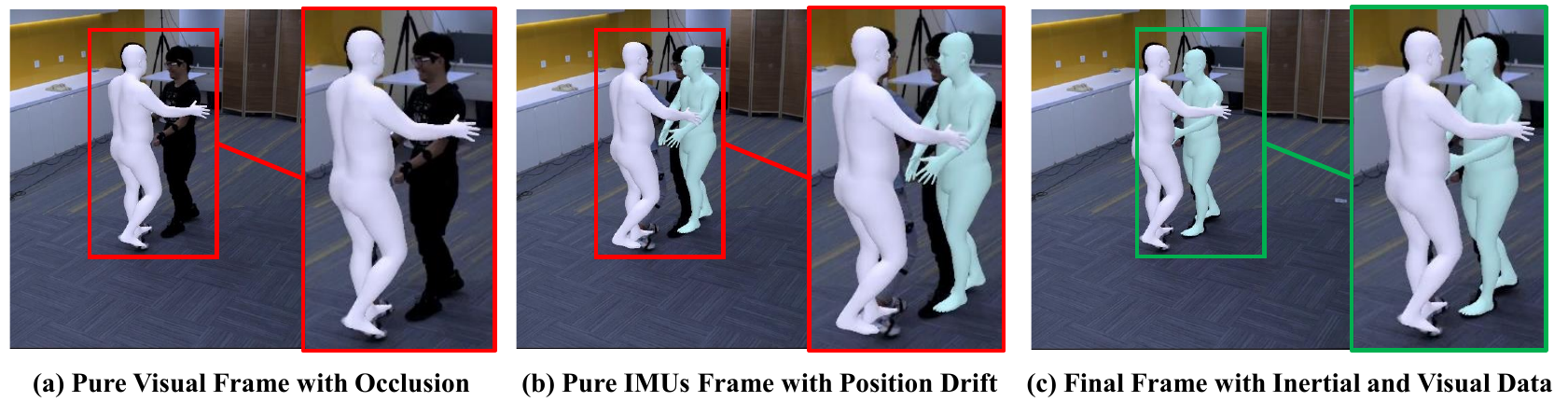}
\caption{\textbf{Example frame of motion recovery with inertial and visual data.}}
\label{fig4c:motionrecovery}
\end{figure}

\begin{figure}[t]
\centering
\includegraphics[width=0.99\linewidth]{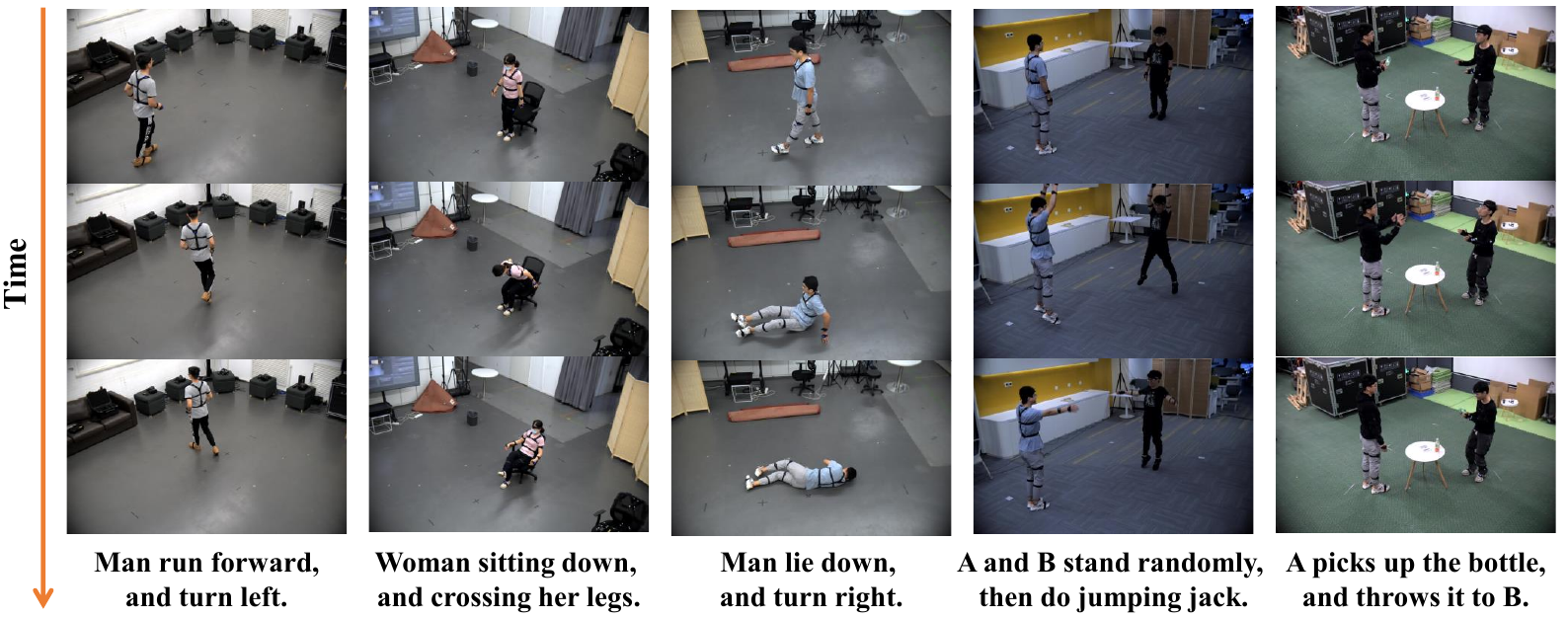}
\caption{\textbf{Fine-grained Actions.} MINIONS contains 121 single-player actions and 25 multi-player actions, including common person-person and person-object interactive actions in daily life.}
\label{fig5:example}
\end{figure}

\textbf{Motion Recovery from Multimedia data.}
Most human motion datasets~\cite{hm36:journals/pami/IonescuPOS14, MPI3DHP:conf/3dim/MehtaRCFSXT17} with a single subject directly recover motion sequences from pure vision sensors.
However, as shown in \figurename~\ref{fig4c:motionrecovery} (a), occlusion between multiple subjects and objects is inevitable, a serious problem for pure visual recovery.
Fortunately, despite the global position drift, we can still obtain accurate motions of actors wearing full-body IMUs under occlusion and varied illumination, as shown in \figurename~\ref{fig4c:motionrecovery} (b).
Therefore, for the joints that cannot be captured accurately by previous visual steps, we fill in the missing joints through full-body IMUs data to obtain the final 3D joints $P_{3d}$ and 3D mesh. 
In detail, for frames $x_i$ whose joints cannot be determined due to occlusions or low illumination, we search simultaneously forward and backward for visual frames $x_{i-n}$ and $x_{i+m}$ with high confidence ($n,m \geq 1$).
Based on the high-confidence joints of $x_{i-n}$ and $x_{i+m}$, and IMUs data such as acceleration and velocity from $x_{i-n}$ to $x_{i+m}$, we estimate the ground truth joint locations of frame $x_i$.
According to 3DPW~\cite{3DPW:conf/eccv/MarcardHBRP18}, the accuracy of IMUs decreases as $m$ and $n$ increase over time. 
However, the values of $m$ and $n$ are typically controllable due to the aid of visual information, thus ensuring the accuracy of our estimation.
After that, we adopt SMPL~\cite{SMPL:journals/tog/LoperM0PB15} as the parameterized representation and fit it to the 2D joints $P_{2d}$, 3D joints $P_{3d}$ and IMUs acceleration, as shown in \figurename~\ref{fig2:datasetall} (d).
We constrain the global motion parameters as follows:
\begin{equation}
   E(\beta,\theta, t) = E_{2d} + E_{3d}  + E_{smooth} +  E_{reg},
\label{equ2:texture}
\end{equation}
where $E_{2d}$ and $E_{3d}$ penalizes the error between the detected 2D/3D joints and the final 2D/3D joints, 
$E_{smooth}$ uses the acceleration information from the IMUs to smooth the final global motion,
$E_{reg}$ is a priori term~\cite{smplify:conf/eccv/BogoKLG0B16} applied to make the results more reasonable.
The $E_{2d}$ and $E_{3d}$ loss is as follows:
\begin{equation}
   E_{2d} = \sum  w_{2d} \cdot  \left \| K \mathcal{J} M(\theta,\beta , t) - P_{2d} \right \|_2^2 ,
\label{equ3:2d}
\end{equation}

\begin{equation}
   E_{3d} = \sum  w_{3d} \cdot  \left \| \mathcal{J} M(\theta,\beta , t) - P_{3d} \right \|_2^2 ,
\label{equ3:3d}
\end{equation}
where $\beta$ denotes the static SMPL shape parameters obtained in~\ref{dataset:textured}, $\theta$, $t$, and $w_{2d}/w_{3d}$ denote the SMPL pose parameters, translation of the camera, and weighted scores, 
$\mathcal{J}$ is a pre-trained linear regression matrix used to generate 3D joints from the SMPL mesh $M(\theta,\beta, t)$, $K$ is the camera intrinsic parameters to transfer 3D joints to 2D joints.

\begin{figure}[t]
\centering
\includegraphics[width=0.95\linewidth]{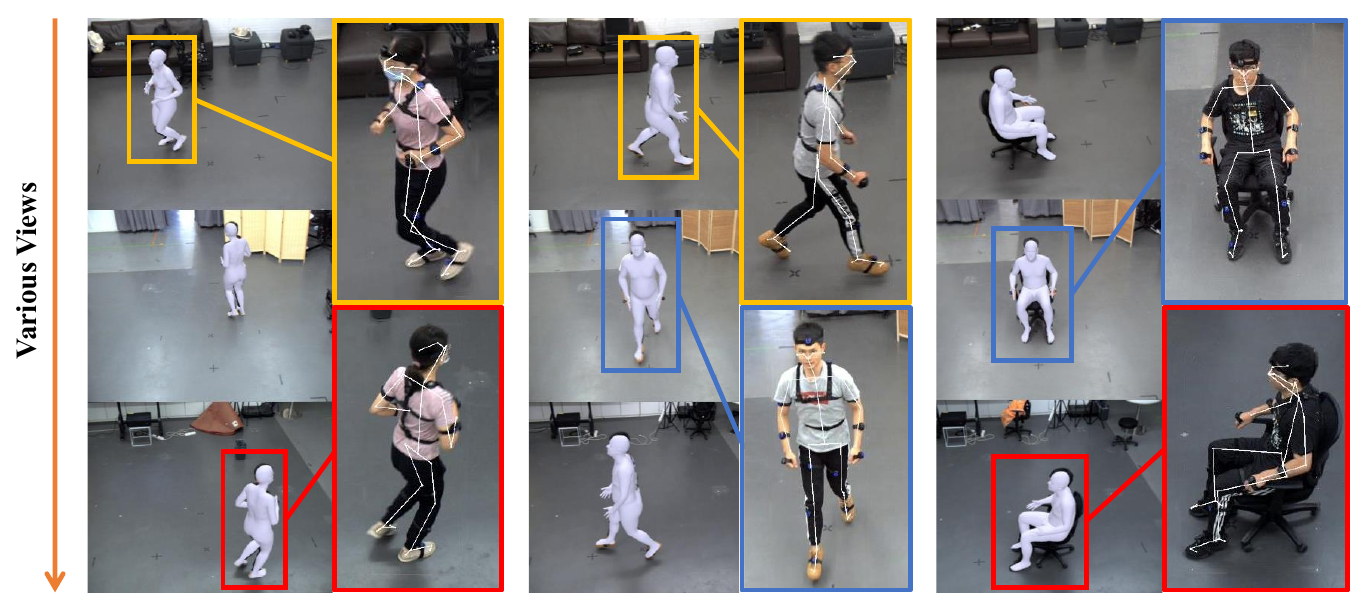}
\caption{\textbf{Qualitative results from single-subject motion capture collection.} Results of 3D mesh and the corresponding re-projected full-body 2D joints from various views, visualized with white rendering for enhanced clarity.}
\label{fig4a:single}
\end{figure}

\begin{figure}[t]
\centering
\includegraphics[width=0.95\linewidth]{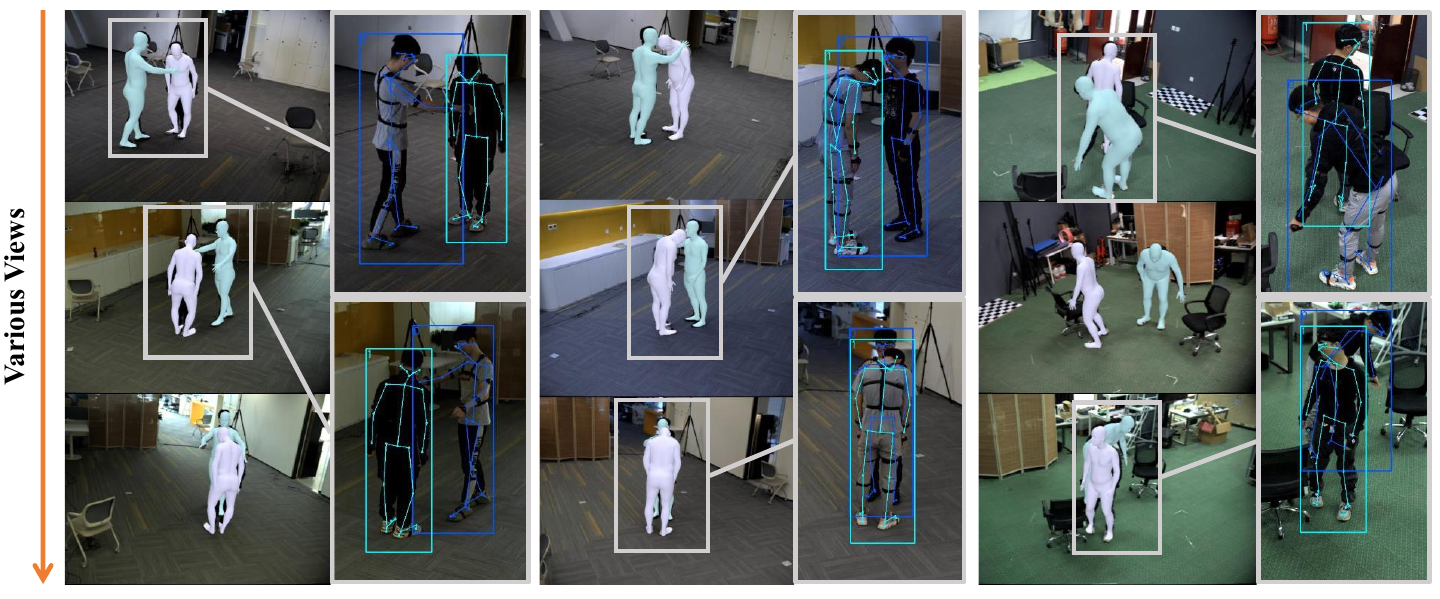}
\caption{\textbf{Qualitative results from multi-subject motion capture collection.} Results of 3D mesh and the corresponding re-projected full-body 2D joints from various views. The identities of subjects are distinguished by colors.}
\label{fig4b:multi}
\end{figure}

\subsection{Dataset Statistics}
\label{subsec:Statistics}

The detailed statistics of the MINIONS are listed in \tablename~\ref{tab1:overview}.
MINIONS is divided into MINIONS-S and MINIONS-M depending on the number of actors.
MINIONS-S is collected for single-actor actions, which contain a total of 4.5 million video frames, 315 minutes of video duration, and 121 categories of fine-grained single actions.
In contrast, MINIONS-M captures the interactive actions between multiple actors, containing a total of one million video frames, 125 minutes of video duration, and 25 categories of common person-person and person-object interactive actions in daily life. 
The detailed action lists are shown in the supplementary materials. 
Note that informed consent was obtained from all individual participants included in the study.

The statistic on the number of frames for different action sequences reflects that most action sequences are smaller than 300 frames (6 seconds), and the longest one can be up to 1500 frames (50 seconds), which shows the complexity of human actions in life. We show some fine-grained action examples of our dataset in \figurename~\ref{fig5:example}.
Moreover, we show the qualitative results of our annotated 3D SMPL mesh and the corresponding re-projected full-body 2D joints from different viewpoints in \figurename~\ref{fig4a:single} and \figurename~\ref{fig4b:multi}.
Benefiting from the multi-view vision sensors and full-body inertial devices, our pipeline can recover the whole-body human motion and global position under various environmental conditions.

\section{Multi-modal Human Motion Capture}

In this section, we present a \textbf{SparseNet} framework to explore the supplementary features from a monocular video and sparse IMUs for lightweight and easy-to-use human motion capture.
Before detailing the network structure, we first introduce the theoretical assumptions in the following section.

\begin{figure*}[t]
\centering
\includegraphics[width=0.95\linewidth]{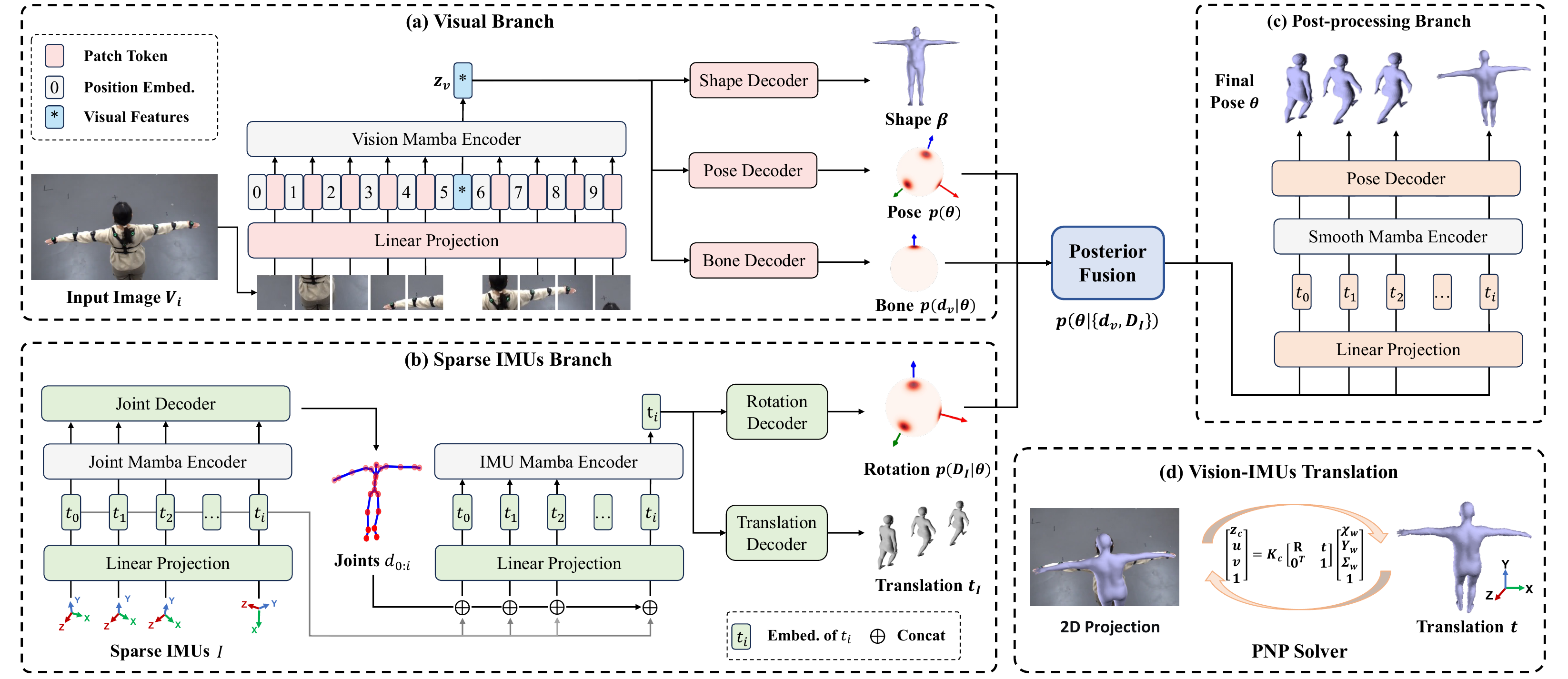}
\caption{\textbf{Overview of the multi-model motion capture framework SparseNet.} (a) The visual branch is designed to regress the body shape parameters $\beta$ and prior pose distribution $p(\theta)$, and the corresponding bone direction distribution $p(d_v | \theta)$.
(b) The Sparse IMUs Branch is designed to infer the comprehensive full-body rotation distribution $p(D_I|\theta)$ from noisy, sparse inertial signals.
(c) The post-processing branch is designed to fusion data from inertial and vision sensors, and smooth the final pose $\theta$.
(d) The global translation $t$ is computed as a PNP-like problem.}
\label{fig3:method}
\vspace{-3mm}
\end{figure*}

\subsection{Theoretical Assumptions}
When both camera and IMUs are available, the estimation of human joint rotation $\theta$ can be formulated as a probabilistic inference problem on the special orthogonal group $SO(3)$ and unit sphere $S^2$. 
To robustly integrate these heterogeneous sensor data and account for their respective uncertainties, we adopt a Bayesian fusion strategy.
Following the previous theoretical foundation~\cite{ta1:downs1972orientation, ta2:khatri1977mises, ta3:fang2023learning}, we model the joint rotation $\theta$ as a latent variable with a prior distribution, and fuse multi-modal observations (visual and inertial).
Specifically, the prior distribution of joint rotation is modeled as a Matrix Fisher distribution over $SO(3)$, denoted as $p(\theta)$. The visual branch estimates 3D bone directions distribution $S^2$ from monocular images, which can be formulated as directional observations following a von Mises-Fisher (VMF) distribution~\cite{ta2:khatri1977mises}:

\begin{equation}
p(d_{v}|\theta) = \frac{1}{c(\kappa_{v})} \exp\left(\kappa_{v} l_{v}^T {\theta}^T d_{v}\right),
\label{eq:dv}
\end{equation}
where $l_{v}$ is the reference bone direction in the T-pose, and $\kappa_{v}$ is the concentration parameter reflecting confidence, $d_{v}={\theta} l_{v}$ is the current bone directional. 
Similarly, the IMU branch provides rotational observations $p(D_{I}|\theta)$.
Assuming conditional independence of observations, the posterior distribution of joint rotation given both modalities is

\begin{equation}
p(\theta|d_{v}, D_{I}) \propto p(\theta) \cdot p(d_{v}|\theta) \cdot p(D_{I}|\theta) .
\label{eq:dvdi}
\end{equation}

This Bayesian formulation allows for flexible multi-sensor fusion, supporting both training and inference stages. It enables robust handling of sensor uncertainty, as the posterior distribution naturally reflects the reliability of each modality.

\subsection{Network Structure}

The overall architecture of SparseNet is shown in \figurename~\ref{fig3:method}. 
To effectively leverage the complementary strengths of vision and inertial sensing, we design a two-branch structure.
The visual and sparse IMUs branch is designed to learn latent features from visual images and sparse inertial sensor signals. 
Then, the posterior fusion module aligns the pose distributions from both branches, resulting in a more accurate and smoother pose rotation distribution.
Finally, we consider the problem of computing translation $t$ from camera parameters $(K, R|T)$, the SMPL~\cite{SMPL:journals/tog/LoperM0PB15} pose rotation $\theta$, and shape $\beta$ as a Perspective-N-Point (PNP) problem and solve it through the PNP solver.

\textbf{Visual Branch} is designed to reconstruct the 3D body shape $\beta$, prior pose distribution $p(\theta)$ and the bone direction distribution $p(d_v | \theta)$ of subjects from monocular RGB frames, as depicted in Fig.~\ref{fig3:method}~(a). 
This branch addresses the limitations of IMU-only systems by providing rich visual context and accurate bone locations, which are crucial for mitigating global drift and recovering human body shapes.
Let $\mathbf{V} = \left \{ V_{i} \right \}_{i=1}^{L}$ denote the input video sequence, where each $V_i \in \mathbb{R}^{3\times H\times W}$ represents the $i$-th RGB frame with spatial resolution $H \times W$, and $L$ is the total number of frames.
The visual branch employs a vision encoder $f_{ve}$, based on the Vision Mamba~\cite{vim} backbone, to extract visual features $z_v$ from the input frames. 
These features are subsequently processed by a lightweight shape decoder $f_{sd}$ and a pose decoder $f_{pd}$ to regress the SMPL~\cite{SMPL:journals/tog/LoperM0PB15} body shape parameters $\beta$ and prior pose distribution $p(\theta)$.
Besides, a bone decoder $f_{bd}$ is used for estimating the bone directions $d_v$ and the corresponding bone direction distribution $p(d_v | \theta)$ is computed according to Eq.~\ref{eq:dv}.
This architectural design enables the visual branch to robustly recover detailed 3D body shape and pose information from single-view RGB inputs, facilitating accurate full-body reconstruction.

\begin{table*}[t]
\small
\caption{Comparison between IMUs-based, monocular vision-based, and multi-modal human motion capture. \textbf{\#IMUs}: the number and placement of IMUs used in the algorithms. 
In detail, \textbf{2} means \{Head, Hip\},
\textbf{4} means \{LeftHand, RightHand, Head, Hip\},
\textbf{6} means \{LeftForeArm, RightForeArm, LeftLeg, RightLeg, Head, Hip\}, 
\textbf{8} means \{LeftHand, RightHand, LeftForeArm, RightForeArm, LeftFoot, RightFoot, Head, Hip\}, 
\textbf{10} means \{LeftHand, RightHand, LeftForeArm, RightForeArm, LeftFoot, RightFoot, LeftLeg, RightLeg, Head, Hip\}.
 \textbf{\#Cams}: the number of camera views used in the algorithms. \textbf{$\mu_{glo}$} and  \textbf{$\sigma_{glo}$}: the mean and variance global rotation error of all body joints in degrees. \textbf{MPJPE}: the mean Euclidean distance between the predicted 3D joint positions and the corresponding ground truth joint positions. \textbf{Jitter}: the average jerk of body joints.}
\centering
\setlength{\tabcolsep}{2mm}{
\begin{tabular}{c|c|c |cc|c|c} 
\midrule[1.5pt]
                                & \#IMUs & \#Cams & $\mu_{glo}$$\downarrow$ [deg] & $\sigma_{glo}$$\downarrow$ [deg] &  MPJPE $\downarrow$ [mm]  & Jitter $\downarrow$ [$10^{2}m/s^{3}$]   \\ 
\midrule[1.5pt]
\multirow{5}{*}{PIP~\cite{yi2022physical}} & 2  & 0  & 24.43 & 17.11 & 128.45 & 8.36 \\
& 4 & 0  & 16.79 & 13.59 & 79.17 & 1.74 \\
& 6  & 0  & 15.73 & 12.53 & 77.23 &  1.95\\
& 8 & 0  & 13.46 & 10.20 & 64.88  &  1.35\\
& 10 & 0  & 13.46 & 10.29 & 64.35 &  1.23\\
\midrule
\multirow{5}{*}{PNP~\cite{pnp:yi2024physical}} & 2  & 0  & 23.93 & 15.82 & 126.21 & 7.99 \\
& 4 & 0  & 15.94 & 12.35 & 74.93 & 1.25 \\
& 6  & 0  & 14.53 & 11.67 & 71.39 & 1.40 \\
& 8 & 0  & 12.52 & 10.38 & 57.75 & 1.69 \\
& 10 & 0  & 12.06 & 10.21 & 55.16 & 1.54 \\
\midrule
\multirow{5}{*}{IMUs-based} & 2  & 0  & 23.62 & 16.91 & 119.56 & 6.01 \\
& 4 & 0  & 14.79 & 12.04 & 68.94 & 1.71 \\
& 6  & 0  & 11.67 & 8.65 & 57.93& 1.17 \\
& 8 & 0  & 11.20 & 8.06 & 53.55& 1.06 \\
& 10 & 0  & 11.09 & 8.31 & 53.20& 1.31 \\

\midrule[1.5pt]
TokenHMR~\cite{dwivedi_cvpr2024_tokenhmr} & 0 & 1 &  12.89 & 6.16 & 49.36 & 10.06 \\ 
\midrule
PromptHMR~\cite{wang2025prompthmr} & 0 & 1 &  11.45 & 8.87 & 47.67 & 11.67 \\ 
\midrule
Vision-based & 0 & 1  & 10.27 & 7.20 & 45.61 & 13.02 \\ 
\midrule[1.5pt]
\multirow{5}{*}{Multi-modal}  & 2  & 1 & 10.18 & 6.96 & 44.80 & 6.24 \\
 &4 & 1 & 9.75 & 6.94 & 41.53 & 1.79 \\
 &\textbf{6} & \textbf{1}&  \textbf{9.20} & \textbf{6.19} &\textbf{39.99}& \textbf{1.57} \\
 &8 & 1&  8.86 & 5.98 &39.39& 1.70 \\
 &10 & 1&  8.81 & 5.95 &39.43& 1.63 \\
\midrule[1.5pt]
\end{tabular}}
\vspace{-5mm}
\label{tab2:experiment}
\end{table*}

\textbf{Sparse IMUs Branch} is designed to infer comprehensive full-body pose information from noisy, sparse inertial signals, as illustrated in Fig.~\ref{fig3:method} (b). 
This branch is crucial for providing high-frequency, occlusion-resistant motion cues that complement the visual modality, especially during fast movements or partial occlusions where visual information may be unreliable.
Formally, we denote the sequence of real sparse inertial input signals as $\mathbf{I} = \left \{ I_{i} \right \}_{i=0}^{L}$, where each $I_{i} \in \mathbb{R}^{NC}$ represents the $i$-th sample containing the global orientation and acceleration of bones in the global coordinate system. 
Here, $N$ is the number of inertial sensors, $C$ is the dimensionality of the inertial data.
Initially, the sparse signals $I_{i}$ are processed by the joint mamba~\cite{mamba2} backbone $f_{jm}$, which predicts the bone positions $d_{0:i}$ according to $d_{0:i} = f_{jm}(I_{0:i})$.
Compared to directly predicting rotation and global translation, introducing the joint mamba backbone helps improve the stability of the model.
Subsequently, leveraging the estimated bone positions $d_{0:i}$, the IMU mamba encoder, translation decoder, and rotation decoder are employed to estimate the global translation $t_I$, joint rotations $D_I$, and the corresponding rotation distribution $p(D_I|\theta)$. Note that the global translation $t_I$ is used for IMUs-based experiments in section~\ref{sec:exper}.

\begin{figure*}[t]
\centering
\includegraphics[width=0.95\linewidth]{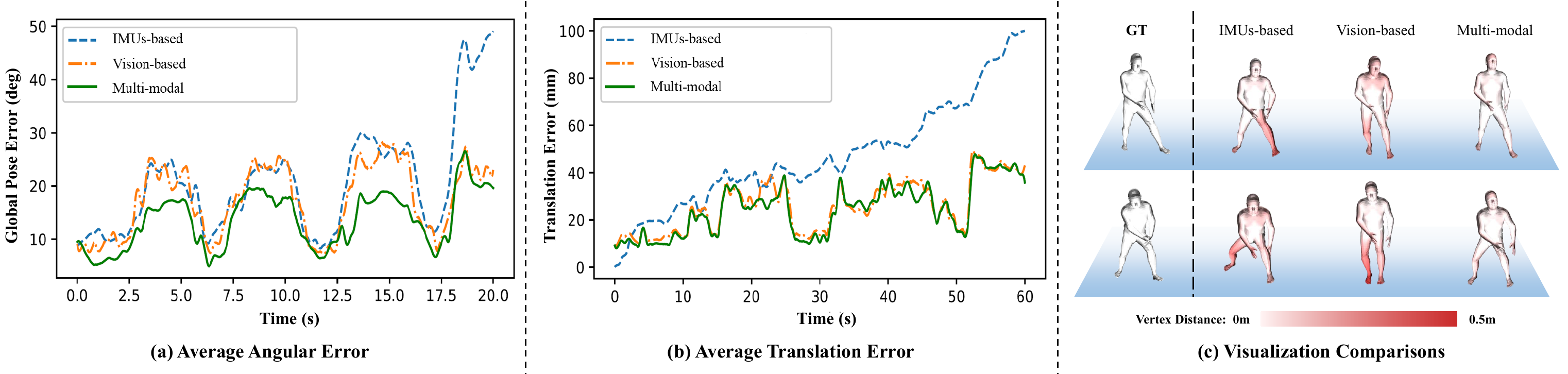}
\caption{\textbf{Visualization. (a): Average angular error (in degrees) over sequences. (b): Average translation error (in mm) over sequences. (c): Visualization comparisons among the IMUs-based, vision-based, and multi-modal human motion capture. The vertices are colored by the distances to the GT.}}
\label{fig7:average}
\vspace{-3mm}
\end{figure*}

\textbf{Post-processing Branch}, as shown in Fig.~\ref{fig3:method}~(c), is designed to integrate the pose distribution $p(\theta)$, the visual bone direction distribution $p(d_v | \theta)$, and the IMUs rotation distribution $p(D_I|\theta)$ from both the visual and IMUs branches, to estimate the fused pose distribution $p(\theta|d_{v}, D_{I})$ as described in Eq.~\ref{eq:dvdi}. 
This fusion module is central to our framework, as it dynamically weights the contributions from each sensor modality based on their estimated uncertainties, thereby achieving more robust and accurate pose estimation than either modality alone.
After that, the smooth mamba backbone is applied for smoothing the final pose $\theta$.
With the fusion pose estimation results, we consider the problem of computing translation $t$ as a PNP-like problem in Fig.~\ref{fig3:method}~(d). 
With the predicted human shape $\beta$, pose $\theta$, 2D joints $P_{2d}$, and given camera parameters $K$, we can easily obtain the global transition $t$ by
\begin{equation}
\arg\min_{t}  \left| {P}_{2d} - K \mathcal{J}  M({\theta},{\beta} , t)  \right|,
\end{equation}

where $K\mathcal{J}$ is a mapping matrix from SMPL mesh $M({\theta},{\beta}, t)$ to 2D keypoints, which is pre-trained in~\cite{SMPL:journals/tog/LoperM0PB15}.

\section{Experiments}
\label{sec:exper}
In this section, we first evaluate the motion capture framework on the MINIONS dataset.
We then conduct ablation studies to evaluate the components of our framework.
At last, we evaluate the methods of various tasks, such as 2D-to-3D keypoint lifting and fine-grained video action recognition on our proposed MINIONS dataset.

\begin{table}[t]
\caption{Results on the TotalCapture dataset for motion capture from a monocular camera and full-body IMUs.}
\label{tab:totalcapture}
\centering
 \setlength{\tabcolsep}{3mm}{
\begin{tabular}{c|c|cc|cc} 
\midrule[1.5pt]
\multirow{2}{*}{Methods} & \multicolumn{2}{c}{TotalCapture~\cite{TotalCapture:conf/bmvc/TrumbleGMHC17}} \\
\cline{2-3}
                         & MPJPE $\downarrow$ (mm)         & PA-MPJPE $\downarrow$ (mm)         \\
\midrule[1.5pt]
DiffCap~\cite{DiffCap:journals/corr/abs-2508-06139}               & 46.2          & 29.9                \\
VIP~\cite{3DPW:conf/eccv/MarcardHBRP18}                      & -             & 26.0             \\
Liu et al.~\cite{Liu:journals/pr/LiuYLZZ24}               & 45.8          & -                \\
\midrule[1.5pt]
Ours &           36.7     &   21.6 \\
\midrule[1.5pt]
\end{tabular}}
\vspace{-3mm}
\end{table}

\subsection{Implementation Details}

We divide the MINIONS dataset into a training set, a validation set, and a testing set by actors. 
The training set contains 12 actors with 3.2 million frames.
The validation set has three actors with 0.9 million frames.
The testing set has five actors with 1.4 million frames.

We train and evaluate the SparseNet framework based on different settings on our MINIONS dataset, including IMU-based, monocular vision-based, and multi-modal human motion capture.
Following previous works, we apply $\mu_{glo}$ (mean global rotation error), $\sigma_{glo}$ (variance of global rotation error), and MPJPE (Euclidean distance error) to measure the mean error between prediction and ground truth in Euclidean space and angular space. 
Additionally, we use the Jitter to measure the average jerk of body joints.

During training, we first pre-train the visual branch with a batch size of 64 for 20 epochs.
The input frames are resized to $512 \times 512$. 
After that, we train the sparse IMUs branch and post-processing branch with a batch size of 512 for 200 epochs using Adam optimizers with a learning rate of 0.001. 
All training and test processes run on an NVIDIA GTX A100.

\subsection{Multi-modal Human Motion Capture}

\begin{table}[t]
\small
\caption{\textbf{FPS comparison across sensing paradigms.} For IMU-based and multi-modal settings, we report FPS under 2/4/6/8/10 IMUs, respectively. ``\#Cams'' denotes the number of cameras.}
\label{tab:fps_paradigm}
\centering
\setlength{\tabcolsep}{3mm}{
\begin{tabular}{l|c|c|c}
\toprule
\textbf{Paradigm} & \textbf{\#IMUs} & \textbf{\#Cams} & \textbf{FPS $\uparrow$} \\
\midrule
\multirow{5}{*}{IMU-based} & 2  & 0 & 522 \\
 & 4  & 0 & 515 \\
 & 6  & 0 & 507 \\
 & 8  & 0 & 501 \\
 & 10 & 0 & 492 \\
\midrule
Vision-based & 0 & 1 & 87 \\
\midrule
\multirow{5}{*}{Multi-modal} & 2  & 1 & 76 \\
 & 4  & 1 & 75 \\
 & 6  & 1 & 75 \\
 & 8  & 1 & 74 \\
 & 10 & 1 & 74 \\
\bottomrule
\end{tabular}}
\end{table}

\textbf{Comparison.}
Our experimental results are detailed in Table~\ref{tab2:experiment}.
The results from our experiments show that motion capture using the monocular camera is frequently hampered by instability resulting from blurriness or occlusion and leads to large Jitter. 
In contrast, motion capture systems that rely on IMUs are prone to larger errors in global rotations and Euclidean distance, attributable to the inherent unreliability of the IMUs. 
However, an integrated system combining just four to six IMUs with a monocular camera can successfully ensure stable motion capture with minimal error and virtually no jitter.
We further evaluate an ultra-sparse setting with only two IMUs (\{Head, Hip\}), which shows a clear performance degradation due to insufficient inertial constraints, but our multi-modal fusion still produces relatively stable motion and suppresses jitter compared to the IMU-only baseline.
In addition, we report the inference speed (FPS) in Table~\ref{tab:fps_paradigm}, showing that our approach runs comfortably above real-time.
Experimental results show that a motion capture system equipped with a monocular camera and more than eight IMUs is unnecessary. They provide negligible improvements in global rotation error and Euclidean distance error, while leading to increased jitter and equipment cost.
Furthermore, we conduct experiments for motion capture from a monocular camera and full-body IMUs on the TotalCapture~\cite{TotalCapture:conf/bmvc/TrumbleGMHC17} dataset in Table~\ref{tab:totalcapture}.
Our method outperforms other methods in terms of performance and demonstrates greater flexibility, allowing for adaptation to various sensor configurations.

\textbf{Computational Cost and Runtime.}
We report the end-to-end inference speed (FPS) for different sensing paradigms. 
All measurements are conducted on a single RTX 4090 GPU. 
As shown in Table~\ref{tab:fps_paradigm}, our method runs at 492--522 FPS in the IMU-based setting and 74--76 FPS in the multi-modal setting (IMU+1 camera), which is well above real-time requirements (e.g., 30 FPS). 
The computational cost is largely independent of action dynamics (slow vs.\ highly dynamic) and scales approximately proportionally with the number of tracked persons.

\textbf{Visualization.}
To facilitate a more intuitive comparison, we provide visualization results of vision-based, IMUs-based, and multi-modal motion capture in Figure~\ref{fig7:average} (c). 
The vertexes of the 3D human body are colored by the distances between the prediction and the ground truth positions.
Additionally, we present the global pose error and positional translation for a motion capture sequence in Figure~\ref{fig7:average} (a) and (b). 
The results reveal that the motion capture based on IMUs tends to exhibit a gradual increase in both global pose error and global positional translation over time. 
However, the integration of a monocular camera and the IMUs can significantly reduce global pose error and also mitigate the inherent global positional drift caused by IMUs.

\subsection{Benchmarks on other Tasks}
In this subject, we evaluate mainstream baselines and benchmarks on tasks including \textbf{2D-to-3D pose estimation}~\cite{simple2dto3d:conf/iccv/MartinezHRL17}, \textbf{Fine-grained Video Action Recognition}~\cite{CSN:conf/iccv/TranWFT19}, \textbf{In-the-wild Monocular Pose Estimation Evaluation}~\cite{3DPW:conf/eccv/MarcardHBRP18} and \textbf{Gait Activity Assessment}, which facilitate more comprehensive explorations of our dataset.

\begin{table}[t]
\caption{2D-to-3D pose estimation. \#Num: Receptive Field (frames). }
\label{supp_tab1:2dto3d}
\centering
 \setlength{\tabcolsep}{2mm}{
\begin{tabular}{c|c|cc|cc} 
\midrule[1.5pt]
\multirow{2}{*}{Method}  & \multirow{2}{*}{\#Num} & \multicolumn{2}{c|}{MPJPE $\downarrow$ (mm)}    & \multicolumn{2}{c}{PA-MPJPE $\downarrow$ (mm)}  \\
\cline{3-6}
                         &                       & Val    & Test                 & Val  & Test                  \\ 
\midrule[1.5pt]
Simple3d~\cite{simple2dto3d:conf/iccv/MartinezHRL17}                 & 1                     & 24.99  & 26.59               & 17.40 & 20.69                \\ 
\hline
\multirow{3}{*}{Dual-Aug~\cite{peng2024dual}} & 27                    & 18.36  & 19.89             & 12.30 & 14.63             \\
                         & 81                    & 17.86 & 19.64             & 11.90 & 14.37               \\
                         & 243                   & 17.18 & 19.22              & 11.40 & 13.95               \\ 
\hline                   
MotionBERT~\cite{Motionbert:zhu2023motionbert} & - &  16.22 & 18.75 & 10.87 & 13.44 \\ 
\midrule[1.5pt] 
\end{tabular}}
\vspace{-3mm}
\end{table}

\textbf{2D-to-3D pose estimation.} 
The 2D-to-3D pose estimation task aims to accurately predict the location of 3D joints from the detected 2D human joints, which is a specific type of 3D pose estimation task.
The challenge is that depth information is lost in the projection from 3D joints to 2D joints, making the reverse process of deducing 3D joints from 2D observations ill-posed.
In \tablename~\ref{supp_tab1:2dto3d}, we evaluate the accuracy of the mainstream methods. 
MPJPE is applied for measuring the average Euclidean distance between the ground truth and predicted joint positions, while PA-MPJPE is a similar metric that first aligns the predicted pose to the ground truth using a rigid transformation before computing the error.
To explore the upper bounds of current mainstream methods, we use the groundtruth of 2D joints as input.
The experimental results indicate that there is still room for improvement in this task.

\textbf{Fine-grained Video Action Recognition.} 
Our dataset provides fine-grained action labels and motion videos from various views, which can be validated for applications in fine-grained video action recognition. 
As shown in \tablename~\ref{supp_tab2:action_recognition}, we test the accuracy of the mainstream video action understanding methods on our dataset.
Following these methods, Top-1 and Top-5 are used to evaluate model performance.
Top-1 is the model's accuracy in predicting the most likely action, while Top-5 is the accuracy at which the true action is within the model's top-five predictions.
Compared to the mainstream video understanding datasets, e.g. Kinetics400~\cite{I3D:conf/cvpr/CarreiraZ17}, NTU RGB+D 60~\cite{ntu60:conf/cvpr/ShahroudyLNW16}, and NTU RGB+D 120~\cite{ntu120:journals/pami/LiuSPWDK20}, our dataset is more challenging since existing methods easily achieve Top-1 accuracy of more than 85\% on the Kinetics400 dataset.

\begin{table}[t]
\caption{Fine-grained Video Action Recognition on datasets.}
\label{supp_tab2:action_recognition}
\centering
\setlength{\tabcolsep}{3mm}{
\begin{tabular}{c|cc|cc} 
\midrule[1.5pt]
\multirow{2}{*}{Method} & \multicolumn{2}{c|}{ MINIONS (Ours) }    & \multicolumn{2}{c}{  Kinetics400~\cite{I3D:conf/cvpr/CarreiraZ17} }  \\ 
\cline{2-5}
                        & Top-1 $\uparrow$    & Top-5 $\uparrow$                &Top-1 $\uparrow$    & Top-5 $\uparrow$              \\ 
\midrule[1.5pt]
TSN~\cite{TSN:journals/pami/0002X00LTG19}                     & 44.94\% & 77.00\%              & 75.89\% & 92.07\%           \\
R(2+1)D~\cite{Rplus1D:conf/cvpr/TranWTRLP18}                & 55.48\% & 85.78\%             & 75.46\% & 92.28\%           \\
SlowFast~\cite{SlowFast:conf/iccv/Feichtenhofer0M19}                & 53.93\% & 83.82\%              & 77.03\% & 92.99\%           \\
TSM~\cite{TSM:conf/iccv/LinGH19}                     & 59.37\% & 88.64\%              & 75.12\% & 91.55\%           \\
TPN~\cite{TPN:conf/cvpr/YangXSDZ20}                     & 64.57\% & 92.11\%              & 76.74\% & 92.57\%           \\
VideoMAE~\cite{VideoMAE:conf/cvpr/WangHZTHWWQ23}  & 73.75\% & 96.01\% & 85.30\% & 96.70\% \\
UniFormerV2~\cite{UniFormerV2:journals/iccv/abs-2211-09552}   & 75.88\% & 96.87\% &  89.50\% & 98.40\% \\
\midrule[1.5pt]         
\end{tabular}}
\vspace{-3mm}
\end{table}

\textbf{In-the-wild Monocular Pose Estimation Evaluation.}
We additionally evaluate our approach on the in-the-wild~\cite{SG1:mihcin2019methodology} monocular benchmark 3DPW~\cite{3DPW:conf/eccv/MarcardHBRP18}. We report the commonly used metrics (all in mm): Procrustes-aligned MPJPE (PA-MPJPE) and MPJPE. As shown in Table~\ref{tab:3dpw_inthewild}, our method achieves 35.4 PA-MPJPE and 56.6 MPJPE, indicating good generalization beyond the controlled capture setup.

\begin{table}[t]
\small
\centering
\setlength{\tabcolsep}{20pt}
\renewcommand{\arraystretch}{1.08}
\caption{\textbf{In-the-wild monocular performance comparison on 3DPW~\cite{3DPW:conf/eccv/MarcardHBRP18}.} We report the commonly used metrics (all in mm): Procrustes-aligned MPJPE (PA-MPJPE) and MPJPE.}
\label{tab:3dpw_inthewild}
\setlength{\tabcolsep}{3mm}{
\begin{tabular}{lcc}
\toprule
Method & PA-MPJPE$\downarrow$ & MPJPE$\downarrow$ \\
\midrule
BEDLAM-CLIFF~\cite{BEDLAM:black2023bedlam} & 46.6 & 72.0 \\
CLIFF~\cite{cliff} & 43.0 & 69.0 \\
HMR2.0~\cite{hmr2} & 44.4 & 69.8 \\
TokenHMR~\cite{dwivedi_cvpr2024_tokenhmr} & 44.3 & 71.0 \\
PromptHMR~\cite{wang2025prompthmr} & 36.6 & 58.7 \\
WHAM~\cite{wham} & 37.5 & 59.8 \\
GVHMR~\cite{shen2024gvhmr} & 37.0 & 56.6 \\
TRAM~\cite{wang2024tram} & 35.6 & 59.3 \\
Ours & 35.4 & 56.6 \\
\bottomrule
\end{tabular}}
\end{table}

\textbf{Part-wise Lower-limb Analysis for Gait Activity Assessment.}
To better understand the gait-related potential~\cite{SG3:mihcin2023database, SG2:mihcin2022simultaneous} of MINIONS, we provide a part-wise analysis on lower-limb components (feet/legs/upper-legs). We report average results over joints belonging to \texttt{LEFTFOOT}, \texttt{RIGHTFOOT}, \texttt{LEFTLEG}, \texttt{RIGHTLEG}, \texttt{LEFTUPLEG}, and \texttt{RIGHTUPLEG}. As shown in Table~\ref{tab:gait_partwise_error}, our multi-modal model consistently improves lower-limb pose accuracy over the IMU-only baseline across all parts.

\begin{table*}[t]
\small
\centering

\setlength{\tabcolsep}{4mm}{
\caption{\textbf{Part-wise lower-limb analysis on MINIONS for gait activity assessment.} We report global orientation error statistics ($\mu_{glo}$/$\sigma_{glo}$, deg), MPJPE (mm), and Jitter ($10^{2}m/s^{3}$), where lower is better. We show results of an IMU-only baseline (\textit{IMUs-based}) and our full multi-modal model.}
\label{tab:gait_partwise_error}
\scalebox{0.9}{\begin{tabular}{c|c|cc|c|c}
\midrule[1.5pt]
Method & Part & $\mu_{glo}$$\downarrow$ [deg] & $\sigma_{glo}$$\downarrow$ [deg] & MPJPE$\downarrow$ [mm] & Jitter$\downarrow$ [$10^{2}m/s^{3}$] \\
\midrule[1.5pt]
\multirow{6}{*}{IMUs-based} & \texttt{LEFTFOOT}   & 13.84 & 10.26 & 62.57 & 1.45 \\
& \texttt{RIGHTFOOT}  & 13.63 & 10.07 & 61.86 & 1.42 \\
& \texttt{LEFTLEG}    & 10.93 & 8.04  & 49.57 & 1.23 \\
& \texttt{RIGHTLEG}   & 10.74 & 7.96  & 48.91 & 1.18 \\
& \texttt{LEFTUPLEG}  & 9.67  & 7.13  & 44.25 & 1.05 \\
& \texttt{RIGHTUPLEG} & 9.58  & 7.09  & 43.82 & 1.03 \\
\midrule[1.5pt]
\multirow{6}{*}{Multi-modal} & \texttt{LEFTFOOT}   & 10.27 & 6.86 & 46.07 & 1.85 \\
& \texttt{RIGHTFOOT}  & 10.08 & 6.67 & 45.43 & 1.82 \\
& \texttt{LEFTLEG}    & 8.47  & 5.78 & 37.25 & 1.64 \\
& \texttt{RIGHTLEG}   & 8.39  & 5.68 & 36.82 & 1.63 \\
& \texttt{LEFTUPLEG}  & 7.64  & 5.16 & 33.57 & 1.48 \\
& \texttt{RIGHTUPLEG} & 7.59  & 5.07 & 33.11 & 1.46 \\
\midrule[1.5pt]
\end{tabular}}}
\end{table*}

\section{Ethical Considerations and Dataset Access}
The principal ethical consideration addressed in this work is privacy, since the dataset contains multi-modal sensor signals and RGB videos that may include personally identifiable appearance information. We take privacy protection seriously and are committed to safeguarding the confidentiality of participants' data.

\textbf{Informed consent.} Informed consent was obtained from all individual participants included in the study. For transparency, the full informed consent form text is provided in the supplementary material.

\textbf{Controlled access.} MINIONS is not intended for indiscriminate public download. The dataset is distributed via a registration-based and approval-based controlled access mechanism. Eligible users from institutions submit the required information and a signed agreement, and then receive download instructions under a Data License Agreement that restricts redistribution and requires responsible use.

\textbf{Project page.} The dataset access instructions are available at: \url{https://huggingface.co/datasets/Sheldong/MINIONS}.

\section{Conclusion}

This paper presents a novel paradigm for consumer-affordable multi-modal motion capture using a monocular camera and sparse IMUs, which offers promising opportunities for personal applications. 
To support research and applications, we have constructed a large-scale motion capture dataset, called MINIONS, using both inertial and vision sensors. The dataset provides a wide range of representations for human motion, including 3D SMPL mesh, 2D/3D joints, and action labels, among others. 
With 5 million frames of 146 fine-grained actions, the dataset is highly scalable for various applications. 
Moreover, we propose a comprehensive analysis to learn discriminative representations from IMUs and cameras by extracting supplementary features.
The results of experiments underscore the distinct benefits offered by the combination of inertial and vision sensors, highlighting the potential of multi-modal motion capture using a monocular camera and sparse IMUs.

\bibliographystyle{splncs04}
\bibliography{main}

\vfill

\end{document}